\documentclass{article}
\usepackage{ijcai25}

\usepackage{times}
\usepackage{soul}
\usepackage{url}
\usepackage[hidelinks]{hyperref}
\usepackage[utf8]{inputenc}
\usepackage[small]{caption}
\usepackage{graphicx}
\usepackage{amsmath}
\usepackage{amsthm}
\usepackage{booktabs}
\usepackage{algorithm}
\usepackage{algorithmic}
\usepackage[switch]{lineno}

\usepackage{multirow}
\usepackage{CJKutf8}
\usepackage{multirow}
\newcommand{\ignore}[1]{}
\usepackage{xcolor}
\usepackage{amsmath}
\usepackage{yhmath}

\title{Hierarchical Multi-Label Generation with Probabilistic Level-Constraint}

\author{
Linqing Chen
\and
Weilei Wang\thanks{Corresponding author.}\and
Wentao Wu\\And
Hanmeng Zhong\\
\affiliations
PatSnap Co., LTD.\\
\emails
\{chenlinqing, wangweilei\}@patsnap.com
}

\begin{document}

\maketitle

\begin{abstract}
Hierarchical Extreme Multi-Label Classification poses greater difficulties compared to traditional multi-label classification because of the intricate hierarchical connections of labels within a domain-specific taxonomy and the substantial number of labels. Some of the prior research endeavors centered on classifying text through several ancillary stages such as the cluster algorithm and multiphase classification. Others made attempts to leverage the assistance of generative methods yet were unable to properly control the output of the generative model.  We redefine the task from hierarchical multi-Label classification to Hierarchical Multi-Label Generation (HMG) and employ a generative framework with Probabilistic Level Constraints (PLC) to generate hierarchical labels within a specific taxonomy that have complex hierarchical relationships. The approach we proposed in this paper enables the framework to generate all relevant labels across levels for each document without relying on preliminary operations like clustering. Meanwhile, it can control the model output precisely in terms of count, length, and level aspects.
Experiments demonstrate that our approach not only achieves a new SOTA performance in the HMG task, but also has a much better performance in constrained the output of model than previous research work.
\end{abstract}

\section{Introduction}

The capacity to proficiently link suitable labels from an extensive label set—which frequently amounts to hundreds of thousands in number—with a provided textual input constitutes a fundamental ability that supports a broad spectrum of applications in multiple domains. This essential skill is notably pertinent in areas such as dynamic search advertising in E-commerce (cited in \cite{Prabhu18,Prabhu14}), open-domain question answering (\cite{Lee19,Chang20}), and taxonomy systems. A practical instance of the latter is the hierarchical categorization employed by online retail giants like eBay and Amazon, which classify products into detailed hierarchies (\cite{Shen20}). This extensive applicability emphasizes the central role of this capability in facilitating refined data handling and informed decision-making across diverse and complex applications.

\ignore{
\begin{figure*}[!t]
\setlength{\abovecaptionskip}{0.2pt}
\begin{center}
\includegraphics[width=4.6in]{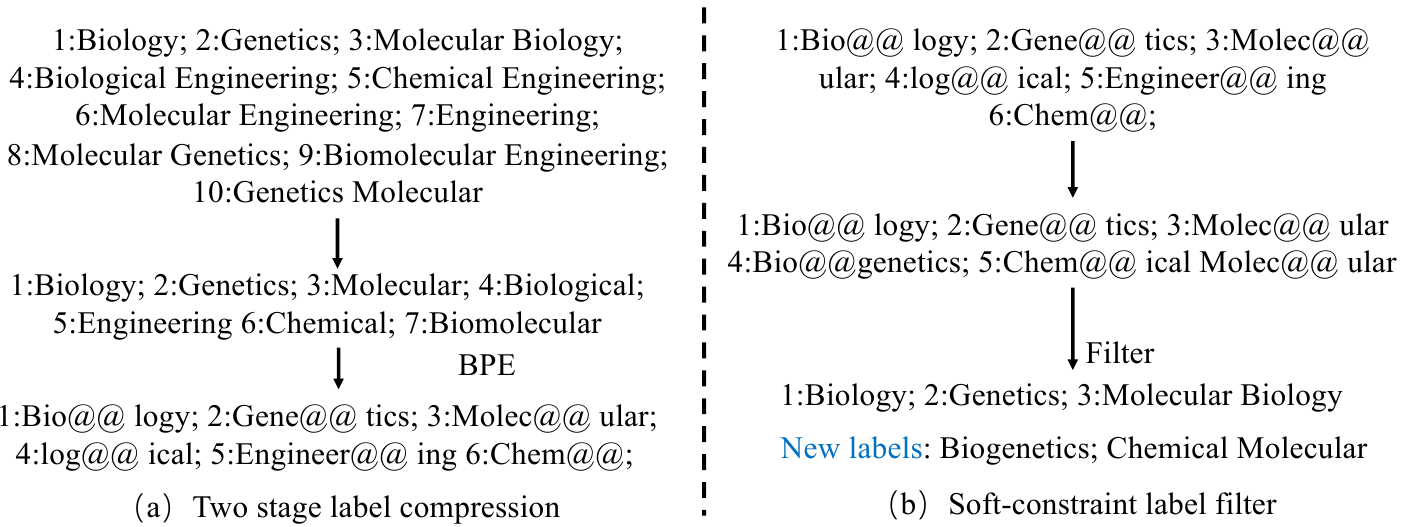}
\end{center}
\caption{(a) Two-stage label compression: multi-word labels are split into single words and further compressed using BPE. (b) Soft-constraint label decoding: real label filters are applied to retain labels from the taxonomy. In this illustration, some filtered-out labels are meaningful but could also be meaningless fragments in practice.} 
\label{fig:label_mapping}
\end{figure*}
}

In contemporary hierarchical multi-label classification methodologies, the predominant approaches typically involve techniques like label vector compression \cite{Liu17}, multifaceted classification paradigms, and complex clustering strategies \cite{Chang20}. Although these strategies have made significant progress in the field, they possess notable drawbacks that demand attention. Specifically, several key issues emerge: 
First, many existing methods neglect the holistic structure inherent in the label system, potentially omitting crucial information. This oversight can impede comprehensive understanding and accurate classification. Second, partitioning the label space into distinct subsets may introduce biases, affecting the fairness and equity of the classification process. Finally, using a path or relationship between labels within a specific taxonomy by hard constraint also involves bias and error propagation, which is disastrous for real-world applications sometimes.
To update and expand existing topic taxonomy frameworks by incorporating novel subjects, studies such as \cite{zhang2018taxogen} and \cite{shang2020nettaxo} utilize clustering mechanisms to facilitate document grouping for informed topic determination. Lee \cite{lee2022taxocom} introduces novelty-adaptive clustering; however, the assignment of labels to emerging topics relies on inheritance from central nodes, which requires further exploration. Notably, Zhang \cite{zhang2021taxonomy} presents a unique approach using a triplet matching network to discern complex relationships between diverse hyponyms, thereby enhancing taxonomic accuracy and fidelity. This combination of strategies highlights  evolving efforts to refine and improve the effectiveness of topic taxonomy systems.

Inspired by the extensive adoption of the Sequence-to-Sequence (Seq2Seq) model—which has been proficiently applied in machine translation as seen in \cite{Bahdanau15,Luong15,Sun17}, summarization as in \cite{Rush15,Lin18}, and style transformation as in \cite{Shen17,Xu18}—we put forward a novel generative approach to tackle the intricate challenge of hierarchical multi-label classification. This innovative strategy is based on the achievements of Seq2Seq models and effectively overcomes the limitations of previous research in this field.


\begin{itemize}

\item We propose an innovative method that combines probabilistic level soft-constraints with label filtering, which ensures that the model's outputs are completely aligned with the label system and respect its hierarchical relationships.

\item For the first time, our generative framework can precisely control the length and level of the model output labels. Thereby, dramatically improve the practicality for real-world application. 

\item We introduce a novel and efficient generative framework that effectively addresses extreme multi-label text classification tasks without the need for external procedures, computations, or additional information.

\end{itemize}

\section{Related Work}

\paragraph{Previous Methods} Liu et al. \cite{Liu05} and Cai et al. \cite{Cai04} transformed the challenge of Hierarchical Extreme Multi-Label Classification (HXMC) into binary classification problems using support vector machines. The SLEEC approach \cite{Bhatia15} addresses the "long-tail distribution" in class labels through the compression of the vector of labels.
Kurata et al. \cite{Kurata16} advocated the use of Convolutional Neural Networks (CNNs) for multi-label classification. Chen et al. \cite{Chen17} combined CNN and Recurrent Neural Networks (RNNs) to capture semantic nuances in labels and text. Liu et al. \cite{Liu17} advanced TextCNN to enhance semantic representation.
Chang \cite{Chang20} explored pre-trained language models for their inherent capabilities. Liu \cite{Liu17} integrated deep learning with sparse embedding strategies to capture extreme multi-label system intricacies.
Yang \cite{Yang18} introduced dual Long-Short-Term Memory (LSTM) to improve the encoding and decoding of textual information, enhancing contextual understanding.
Wang \cite{Wang21} proposed partitioning the class hierarchy into discrete levels to refine the treatment of hierarchical relationships in classification.

\paragraph{Generative Methods} Jung et al. \cite{jung2023cluster} introduce XLGen, a generative approach utilizing pre-trained text-to-text models enhanced by cluster guidance to manage tail labels and semantic relations. However, its generative nature may introduce noise through irrelevant labels and reliance on cluster guidance could limit flexibility across diverse datasets.
Xiong \cite{xiong2023xrr} proposes a novel two-stage XMTC framework with candidate retrieval and deep ranking (XRR). Despite its innovation, computational complexity in these stages may become a bottleneck when applied to large-scale datasets.
To our knowledge, Chen et al.\cite{Chen2022GenerationMF} first proposed applying the transformer framework for extreme multi-label classification task.
Ostapuk \cite{ostapuk2024follow} presents HECTOR, a method for extreme multi-label classification that uses Transformer architecture to predict paths within a label hierarchy. This approach may suffer from error propagation during path prediction, and its dependence on hierarchical structures could reduce effectiveness with non-hierarchical data.

\section{Generation Hier-Multi-Label with Probabilistic Constraint}

\begin{figure}[!h]
\begin{center}
\resizebox{\linewidth}{!}{
\includegraphics[width=4.6in]{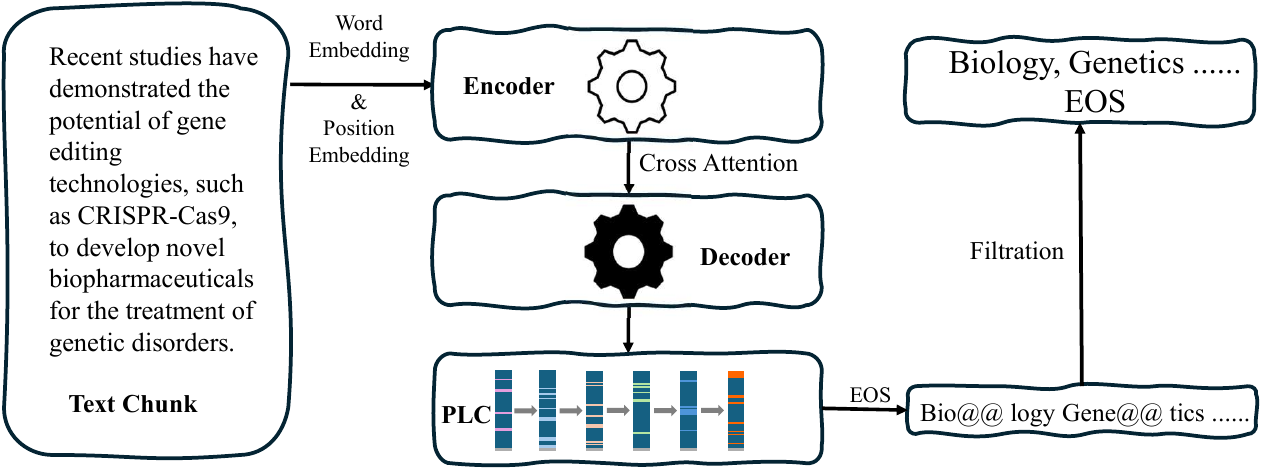}
}
\end{center}
\caption{Overall illustration of our proposed model.} 
\label{fig:architecture}
\end{figure}

We formulate the Hierarchical Multi-label Generation task as follows:

Given a label space $\mathcal{L} = \{l_1, l_2, \cdots, l_L\}$ and a text sequence $\mathcal{X} = \{x_1, x_2, \cdots, x_m\}$, the task is to identify the most correlated subset $\mathcal{Y} = \{l_1, l_2, \cdots, l_n\}$ of $\mathcal{L}$, calculated as:

\begin{equation}
    p(\mathcal{Y}|\mathcal{X}) = \prod_{i=1}^{n} P(l_i | l_1, l_2, \cdots, l_{i-1}, \mathcal{X}),
\end{equation}

The framework is based on an encoder-decoder model that integrates a probability based soft constraint mechanism and a label filtering component, as illustrated in Figure~\ref{fig:architecture}.

Our proposed procedure of text chunk tagging, or we call it HMG is quite efficient and effective. 1) Embedding the text chunk that waits to tagging; 2) Encoding the embedded text vector and deliver it to the decoder for cross attention mechanism; 3) Decoding the output token based on the previous output or the begin of sequence special token (BOS) and output of encoder. 4) output the labels which are precisely controlled by generation hyper parameters; 5) filtering the repeat labels and the labels do not exist within taxonomy.

\ignore{\textcolor{blue}{Describe the framework step by step with a fancy pic.}}

\subsection{Text Encoding with Multiple Heads}


The input text is formed by concatenating the title and abstract of papers or patents, similar to the method used by Press \cite{Press16}. The model's encoder and decoder share a linear transformation matrix. We use two separate position encoding functions to encode the title and abstract:

{\small{
\begin{equation}
    I = \text{Concat}(PE_1 + \text{emb}(T), PE_2 + \text{emb}(A)),
\end{equation}
}}

where $I$ represents the input sequence with added positional information, $T$ and $A$ denote the title and abstract, $emb()$ indicates word embedding, and $PE()$ denotes position encoding.

\subsection{Probabilistic Level Soft-Constraint}

To minimize the decoder-side vocabulary, we resort to Byte Pair Encoding (BPE) as delineated in Sennrich et al.~\cite{Sennrich16}. \ignore{ As can be seen in Figure~\ref{fig:label_mapping}(a), }The adoption of BPE significantly reduces the quantity of sub-wordized category labels from approximately 700K to a manageable number below 30K. To efficiently generate labels, we utilize Beam Search (BS). Notably, the use of BS ensures that the time complexity of predictions depends not on the size of the label set $\mathcal{L}$, but solely on the beam dimensions and the average length of label representations, due to our use of auto-regressive generation.

\begin{figure}[!ht]
\setlength{\abovecaptionskip}{0pt}
\begin{center}
\includegraphics[width=1.6in]{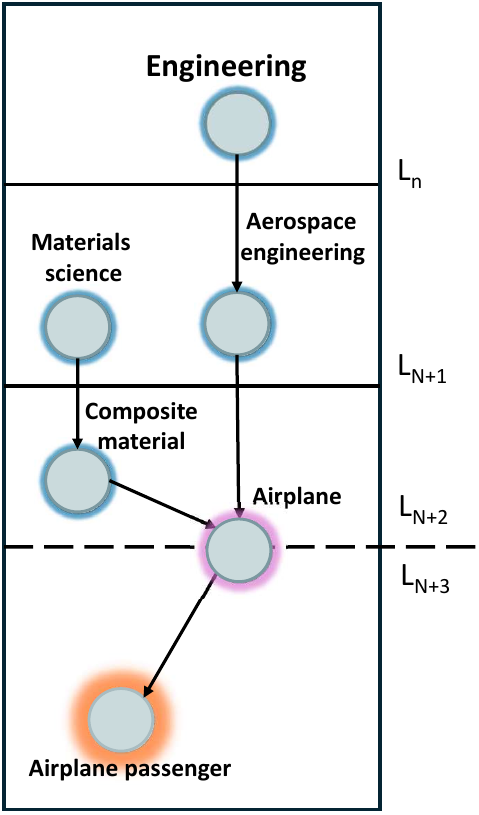}
\end{center}
\caption{The illustration of the complex cross level relationship within taxonomy and the error within it.} 
\label{fig:label_hier}
\end{figure}

\paragraph{Motivation of PLC}
Firstly, methods proposed by \cite{ostapuk2024follow} have two un-property aspect (1) This method could not precisely control the output of the model, such as the level and the count of the model output label. In many real-world applications of multi-label classification/generation, there is no need to extend the labels into thousands. Such as MAG taxonomy, which try to classify the paper and patent into record science field. (2) As Fig.~\ref{fig:label_hier} had illustrated above, there are many errors in the path of taxonomy like MAG. Using the path between labels is not a good way to predict the label.

Additionally, the method proposed by Chen et al.~\cite{Chen2022GenerationMF} though tried to control the output labels of model, but roughly rely on the transformer framework itself. In our opinion, it's too soft and too uncontrollable. Building on this observation, we propose the use of PLC, moving away from the traditional hard constraints of earlier studies.

\paragraph{Preparation for PLC}
During each decoding operation, the decoder yields an output vector that matches the length of the vocabulary in terms of dimensionality. After performing a softmax operation, a token is selected corresponding to the dimension with the highest probability. Motivated by this paradigm, we decode the vocabulary to observe the distribution of tokens that constitute labels at distinct hierarchical levels.

\begin{figure}[!t]
\begin{center}
\includegraphics[width=2.0in]{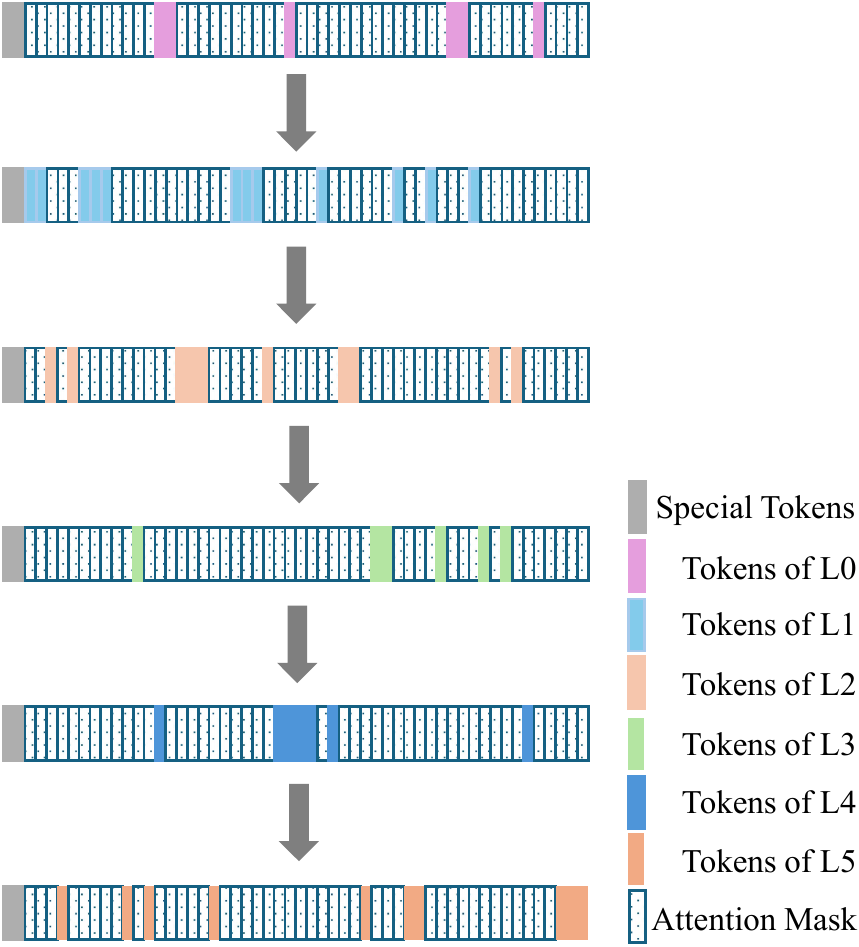}
\end{center}
\vspace{-10pt}
\caption{Overall illustration of our proposed probability-based soft-constraint method.} 
\label{fig:soft-constraint}
\end{figure}

\paragraph{Construction of PLC}
As illustrated in Figure~\ref{fig:soft-constraint}, the level-0 label attention mask eliminates all dimensions in the output vector that correspond to non-level-0 label tokens. This ensures these dimensions do not participate in the softmax computation, thus preventing the model from producing any tokens unrelated to level-0 labels. Importantly, each hierarchical level's mask retains special tokens such as BOS/EOS, thus preserving the model's ability to autonomously conclude its output.

Ultimately, during the decoding phase of the model, the output vector is subjected to a sequence of softmax computations using label token attention masks across different levels. This produces probability vectors associated with various hierarchical levels. Our approach, which employs probability-based soft constraints, ensures that output tokens categorically belong to the corresponding levels without imposing inflexible constraints on the model.

As a result, we obtain attention masks for labels across six hierarchical levels. It is worth noting that since the probability vector needs to be computed only once, our method does not significantly increase the computational load during model inference.

\subsection{Hierarchical Extreme Multi-label Generation}
(1) The framework generates labels of a specific level with our proposed PLC mechanism until the EOS special token is generated. (2) Using the labels from the upper level as context to generate next level labels. Because we want to make the model to learn about how to use the context, we use a little adapter to fuse the upper level labels' vector into text chunk vector during the cross attention procedure.

\subsection{Filtering, Discovering and Completing}

Furthermore, in order to generate only labels that exist within the taxonomy label set $\mathcal{L}$, we employ a stringent real label filter, as shown in Figure~\ref{fig:architecture}. This ensures that the output of our model aligns with the predefined taxonomy system, thus ensure the accuracy of our approach.

Labels that are not part of the existing taxonomy label set are identified as new topics. We compare the embedded vectors of these new topics with existing labels in the system and retrieving the most similar vectors as the parent label. Then we compare the new topic with the upper-level and lower-level topics of the parent label, respectively. This comparison helps the framework determine the new topic belongs to which level, allowing us to integrate it into the label hierarchy.

\section{Inadequacy of Large Language Models}

Zhong et al. \cite{zhong2023can} compare ChatGPT with fine-tuned BERT models on NLU tasks. Their findings indicate that while ChatGPT excels in inference tasks, it underperforms on paraphrase and similarity tasks. Despite improvements from advanced prompting strategies, ChatGPT continues to struggle with fine-grained semantic extraction, particularly with negative samples.

As seasoned researchers in the field of generative models, we have long considered the application of large language models (LLMs) for extreme multi-label text generation. In fact, we developed a domain-specific LLM with approximately 70 billion parameters for NLP tasks like retrieval augmented QA. However, we found that using LLMs for HMG is not only cost prohibitive but also presents significant technical challenges that hinder practical implementation.

\begin{itemize}

  \item LLMs have pre-trained on vast datasets, with complex vocabularies covering multiple languages and over hundreds of thousands of tokens. This complexity can lead to uncontrollable generation results, where output labels are not within the taxonomy system.

  \item Even if using system prompts or few-shot learning, LLms requires inputting hundreds of thousands of tokens for label names and millions for label system relationships as query prefixes, which is impractical.

  \item In real-world applications, the inference speed of LLMs is too slow for tasks requiring the classification of billions of long texts, such as patents and academic papers.

  \item Additionally, the general-purpose nature of LLMs can lead to confusion and inconsistent outputs when dealing with highly specialized and nuanced text tagging.\ignore{\textcolor{red}{Maybe, we can add a case at fig2 (b) to visionlize the shortcoming of LLMs in HMG task.}}

  \ignore{\item Previous research work could not utilize the hierarchical relationship information between labels within the taxonomy.}
\end{itemize}

\section{Experiments}

In order to efficiently compare our framework with previous research work, we utilize the same dataset as that in \cite{ostapuk2024follow} as our main experimental dataset. The results show that our methods have achieved a new SOTA performance. We also demonstrate our framework's cutting-edge capability of precisely controlling model output in the section~\ref{sec:ablation}.

\subsection{Dataset}

The Microsoft Academic Graph (MAG)\footnote{\url{https://www.microsoft.com/en-us/research/project/microsoft-academic-graph}} is a heterogeneous, open-source graph comprising scientific publications. In this work, we use a subset of the MAG dataset. MAG-CS. The Microsoft Academic Graph (MAG) Computer Science (CS) constitutes a subset of the MAG dataset [31], concentrating on the computer science domain. It encompasses papers that were published at 105 top CS conferences spanning from 1990 to 2020. Additionally, the label tree contains relevant concepts that are descendants of the root-level “Computer Science”.

\subsection{Evaluation}
Following previous work Yuan~\cite{Wang21}, we use Micro-F$_1$\cite{Manning08}~ as the main evaluation metric.
Micro-precision (Micro-P) and Micro-recall (Micro-R) are also reported as supplementary references.

\subsection{Settings}

Our model is implemented based on THUMT\footnote{\url{https://github.com/THUNLP-MT/THUMT}}\cite{Tan20}. The dimension of the hidden state vectors is set to 768, and the batch size is set to 40280 characters. The remaining experimental settings remain consistent with the Transformer model.

\subsection{Baselines}

BERT~\cite{Devlin18} is a pre-trained bidirectional Transformer model.
SGM~\cite{Yang18} is a sequence-to-sequence model based on LSTM.
HSG~\cite{Wang21} incorporates weakly supervised semantic guidance.
XML-CNN~\cite{liu2017deep} utilizes convolutional networks with dynamic pooling to map document representations to label space.
AttentionXML~\cite{you2019attentionxml} builds a probabilistic label tree (PLT) and applies multi-label attention at each level.
MATCH~\cite{zhang2021match} leverages document metadata and label hierarchies for extreme multi-label classification.
XR-Transformer~\cite{zhang2021fast} fine-tunes pre-trained Transformers on progressively harder tasks using a hierarchical label tree.
REASSIGN~\cite{romero2023leveraging} aggregates label probabilities along hierarchical paths, selecting the highest-scoring paths.
HECTOR~\cite{ostapuk2024follow} uses hard label relationships to improve prediction accuracy.

\subsection{Experimental Results}
\label{results}


\begin{table*}[!ht]
\centering
\resizebox{\textwidth}{!}{
\begin{tabular}{c|l|ccccccccccc}
\hline
L & Algorithms & P@1 & P@3 & P@5 & N@3 & N@5 & $\mu$Prec & $\mu$Recall & $\mu$F1 & MPrec & MRecall & MF1 \\
\hline
\multirow{7}{*}{2} & XML-CNN & 0.7002 & 0.4516 & 0.3283 & 0.6366 & 0.6390 & 0.6658 & 0.4712 & 0.5518 & 0.2639 & 0.1570 & 0.1969 \\
& AttentionXML & 0.8665 & 0.5884 & 0.4245 & 0.8381 & 0.8406 & \textbf{0.7965} & 0.7125 & 0.7522 & 0.4073 & 0.4732 & 0.4378 \\
& MATCH & 0.8434 & 0.5363 & 0.3763 & 0.7795 & 0.7721 & 0.7826 & 0.5895 & 0.6724 & 0.3989 & 0.3107 & 0.3494 \\
& XR-Transformer & 0.8027 & 0.5437 & 0.3958 & 0.7677 & 0.7717 & 0.7325 & 0.6350 & 0.6803 & 0.3926 & 0.4417 & 0.4157 \\
& REASSIGN & 0.6680 & 0.4224 & 0.3017 & 0.5942 & 0.5901 & 0.7054 & 0.4300 & 0.5343 & 0.0709 & 0.0851 & 0.0773 \\
& HECTOR & 0.8917 & 0.5931 & \textbf{0.4249} & 0.8530 & 0.8527 & 0.7745 & 0.7113 & 0.7416 & 0.3397 & 0.4936 & 0.4025 \\
& OURS & \textbf{0.9021} & \textbf{0.6002} & 0.4241 & \textbf{0.8642} & \textbf{0.8639} & 0.7953 & \textbf{0.7318} & \textbf{0.7622} & \textbf{0.4185} & \textbf{0.5124} & \textbf{0.4602} \\
\hline
\multirow{7}{*}{3} & XML-CNN & 0.6747 & 0.4121 & 0.2931 & 0.6681 & 0.6913 & 0.6115 & 0.5106 & 0.5565 & 0.3213 & 0.1781 & 0.2291 \\
& AttentionXML & 0.8346 & 0.4973 & 0.3440 & 0.8290 & 0.8448 & \textbf{0.8042} & 0.6674 & 0.7294 & 0.4362 & 0.4693 & 0.4521 \\
& MATCH & 0.7818 & 0.4496 & 0.3097 & 0.7583 & 0.7725 & 0.7381 & 0.5909 & 0.6563 & 0.3979 & 0.3658 & 0.3812 \\
& XR-Transformer & 0.7906 & 0.4770 & 0.3297 & 0.7879 & 0.8015 & 0.7432 & 0.6417 & 0.6887 & 0.4540 & 0.4927 & 0.4726 \\
& REASSIGN & 0.6019 & 0.3636 & 0.2574 & 0.5836 & 0.6025 & 0.6879 & 0.4027 & 0.5080 & 0.0846 & 0.1241 & 0.1006 \\
& HECTOR & 0.8918 & 0.5141 & 0.3521 & 0.8745 & 0.8885 & 0.7513 & 0.7181 & 0.7343 & 0.4106 & 0.6507 & 0.5035 \\
& OURS & \textbf{0.9103} & \textbf{0.5324} & \textbf{0.3697} & \textbf{0.8920} & \textbf{0.9063} & 0.7953 & \textbf{0.7396} & \textbf{0.7661} & \textbf{0.4672} & \textbf{0.6832} & \textbf{0.5537} \\
\hline
\multirow{7}{*}{4} & XML-CNN & 0.6662 & 0.3777 & 0.2555 & 0.7358 & 0.7724 & 0.5899 & 0.4959 & 0.5388 & 0.4167 & 0.2129 & 0.2818 \\
& AttentionXML & 0.8113 & 0.4257 & 0.2748 & 0.8581 & 0.8788 & 0.7311 & 0.6297 & 0.6766 & 0.4691 & 0.5711 & 0.5151 \\
& MATCH & 0.7330 & 0.3843 & 0.2547 & 0.7789 & 0.8071 & 0.6675 & 0.5491 & 0.6025 & 0.3876 & 0.4731 & 0.4261 \\
& XR-Transformer & 0.7775 & 0.4083 & 0.2607 & 0.8197 & 0.8364 & 0.7053 & 0.6620 & 0.6830 & \textbf{0.5378} & 0.5499 & 0.5438 \\
& REASSIGN & 0.5416 & 0.3174 & 0.2250 & 0.6015 & 0.6478 & 0.4013 & 0.4879 & 0.4404 & 0.1341 & 0.2735 & 0.1799 \\
& HECTOR & 0.8494 & 0.4390 & 0.2814 & 0.8961 & 0.9140 & 0.7084 & 0.7567 & 0.7317 & 0.5217 & \textbf{0.7197} & \textbf{0.6049} \\
& OURS & \textbf{0.8669} & \textbf{0.4522} & \textbf{0.2955} & \textbf{0.9140} & \textbf{0.9323} & \textbf{0.7397} & \textbf{0.7794} & \textbf{0.7590} & 0.5346 & 0.7157 & 0.6121 \\
\hline
\multirow{7}{*}{5} & XML-CNN & 0.7815 & 0.3376 & 0.2126 & 0.8581 & 0.8736 & \textbf{0.8454} & 0.4014 & 0.5443 & 0.3845 & 0.1346 & 0.1994 \\
& AttentionXML & 0.8612 & 0.3492 & 0.2162 & 0.9101 & 0.9209 & 0.7526 & 0.7534 & 0.7530 & 0.5181 & 0.5013 & 0.5096 \\
& MATCH & 0.7802 & 0.3256 & 0.2080 & 0.8368 & 0.8585 & 0.8370 & 0.5298 & 0.6489 & \textbf{0.5886} & 0.4099 & 0.4832 \\
& XR-Transformer & 0.8213 & 0.3243 & 0.2015 & 0.8551 & 0.8664 & 0.7735 & 0.7087 & 0.7397 & 0.5512 & 0.5733 & 0.5621 \\
& REASSIGN & 0.7121 & 0.3205 & 0.2067 & 0.8022 & 0.8283 & 0.5365 & 0.6067 & 0.5694 & 0.2213 & 0.2466 & 0.2333 \\
& HECTOR & 0.8946 & 0.3526 & 0.2170 & 0.9292 & 0.9370 & 0.7675 & 0.8028 & 0.7848 & 0.5704 & 0.7597 & 0.6516 \\
& OURS & \textbf{0.9125} & \textbf{0.3702} & \textbf{0.2279} & \textbf{0.9478} & \textbf{0.9557} & 0.8382 & \textbf{0.8229} & \textbf{0.8305} & 0.5789 & \textbf{0.7977} & \textbf{0.6707} \\
\hline
\end{tabular}
}
\caption{Performance comparison of our method and other competing methods on HXMC task on MAG-CS dataset.}
\label{table:performance_comparison}
\end{table*}

\paragraph{Comprehensive Comparison on MAG-CS}

Table~\ref{table:performance_comparison} provides a comprehensive comparison of our method against several competing models on the HMG task using the MAG-CS dataset across different hierarchical layers (L). For each method, we report various evaluation metrics, including precision (P@1, P@3, P@5), normalized precision (N@3, N@5), micro-averaged precision, recall, and F1-score ($\mu$Prec, $\mu$Recall, $\mu$F1), as well as macro-averaged metrics (MPrec, MRecall, MF1).
Across all layers, namely L equals 2, 3, 4, and 5, our proposed method unfailingly surpasses the baselines in the majority of metrics. Particularly noteworthy is that the improvement is especially remarkable in micro-averaged metrics, which serve as indicators of overall performance across all classes.
For example, when L is equal to 2, our method attains a $\mu$F1 score of 0.7622. This score surpasses that of the second-best model, AttentionXML, which has a score of 0.7522. Likewise, at L = 5, our method achieves the highest $\mu$F1 score of 0.8305. The closest competitor, HECTOR, scores 0.7848.
Moreover, our approach showcases remarkable performance in precision at rank 1 (P@1). It attains the highest scores in all layers, signifying its potency in ranking the most pertinent labels. For instance, when L equals 5, our method accomplishes a P@1 of 0.9125, surpassing the next most excellent model, HECTOR, which has a score of 0.8946.

In summary, the method that has been proposed not only consistently surpasses strong baselines such as AttentionXML, MATCH, XR-Transformer, and HECTOR, but also shows robustness across diverse metrics, especially in micro-averaged precision, recall, and F1 scores. These results evidently prove the effectiveness of our approach in hierarchical multi-label generation tasks.

\section{Ablation}
\label{sec:ablation}

In this subsection, we use original MAG dataset as experimental dataset.
Table~\ref{tbl:dataset} provides key statistics about the patent and paper datasets employed in our experiments, which are sampled from original MAG dataset. The abbreviation "M" denotes million, and the datasets were obtained via random sampling. 

\begin{table}[!ht]
\small
\centering
\resizebox{\linewidth}{!}{
\begin{tabular}{l|ccc|ccc}
\hline
\multirow{2}{*}{\bf Dataset} & \multicolumn{3}{c}{\bf Paper} & \multicolumn{3}{c}{\bf Patent} \\
\cline{2-7}
&  Train  &  Dev  &  Test  &  Train  &  Dev  &  Test  \\
\hline
Samples & 1.97M & 5,000 & 5,000 & 3.73M & 5,000 & 5,000 \\ 
Labels & 16.90M & 42,966 & 43,206 & 22.98M & 30,797 & 30,623 \\
{\small Avg. Tokens} & 188.55 & 190.14 & 190.34 & 129.81 & 131.38 & 130.91 \\ 
{\small Avg. Labels} & 8.54 & 8.59 & 8.64 & 6.15 & 6.16 & 6.12 \\ 
\hline
\end{tabular}
}
\caption{Key statistics about the experimental datasets.}
\label{tbl:dataset}
\end{table}

\begin{table*}[!t]
\centering
\resizebox{\textwidth}{!}{
\begin{tabular}{l|ccccc|ccccc}
\hline
\multirow{3}{*}{\bf Model} & \multicolumn{10}{c}{\bf Mixed Data} \\
\cline{2-11}
& \multicolumn{5}{c|}{\bf 0-1} & \multicolumn{5}{c}{\bf 0-5} \\
& m-F & m-P & m-R & Rouge-2 & Rouge-L & m-F & m-P & m-R & Rouge-2 & Rouge-L  \\
\hline
\#1~Ours & \bf 79.48 &  83.68 &  75.68 & \bf73.20 & 77.47 & \bf76.10 &  78.00 & \bf74.29 & \bf51.10 & \bf64.96 \\
\hline
\#2~~SGM & 66.14 & 98.08 & 49.89 & 47,25 & 50.13 & 61.25 & 78.56 & 50.19 & 48.46 & 50.89 \\
\#3~$~~$ + Global Emb & 67.02 & 96.74 & 51.27 & 48.96 & 52.10 & 61.99 & 79.67 & 50.73 & 47.39 & 52.46 \\
\#4~$~~$ + PLC & 68.37 & \bf 96.36 & 52.98 & 48.07 & 53.74 & 63.43 & \bf81.60 & 51.88 & 32.65 & 52.30 \\
\#5~~BERT & 75.03 & 77.94 & 72,33 & 71.99 & 74.69 & —— & —— & —— & —— & ——  \\
\#6~$~~$ + PLC & 78.11 & 78.98 & \bf77.26 & 75.88 & \bf79.35 & —— & —— & —— & —— & ——  \\
\hline
\end{tabular}
}
\caption{
Mixed Data MicroF$_1$(\%). \textbf{G-HXMC} refers to the generative method proposed in this paper. \textbf{Soft-Con} indicates the previously discussed soft constraint method. In the \textbf{SGM} results, \textbf{Global Embedding} describes the technique used to preserve LSTM information by utilizing decoding results from prior time steps. \ignore{\textbf{Soft-Info} explores the impact of processing only the target label sequence in training data on enhancing BERT's performance through structural label information.}}
\label{tbl:main_results_mix_data}
\end{table*}

We have modified the SGM to incorporate the probability based soft constraint, which we proposed in this paper and present the experimental results, to verify the effectiveness of the PLC mechanism. Since BERT lacks a decoder architecture, it is not suited for directly selecting highly relevant results from a large label set consisting of hundreds of thousands of options. As a result, the paper does not report performance for layers 0–5. Instead, the evaluation on BERT is focused on layers 0–1, where the label set is reduced to approximately 200. Meanwhile, the layers 2–5 contain almost 700 thousand labels.

\subsection{Performance on MAG-Mixed}

Table~\ref{tbl:main_results_mix_data} presents the performance of the proposed methods on the MAG-Mixed dataset, which includes both papers and patents. The left side presents the classification results for layers ranging from 0 to 1. On the other hand, the right side encompasses the classification results for layers spanning from 0 to 5. 

The results in \#4, and \#6 show that the PLC mechanism brings about significant enhancements for all baseline models. In particular, the method proposed in this paper achieves approximately 10-point gains over SGM in both the 0–1 and 0–5 layer tagging tasks. In the HMG task on 0–1 layers, our approach achieves performance on a par with BERT, which has been pre-trained on a massive corpus. \ignore{When comparing \#1 and \#2, it can be observed that the PLC mechanism introduced herein significantly improves model performance.} Similarly, when comparing \#2 with \#4 and also comparing \#5 with \#6, it becomes evident that the PLC mechanism not only brings improvements to the proposed framework but also enhances the performance of the benchmark model.

\subsection{Performance of Outputs Controlling}

As previously discussed, the control of the level and quantity of output labels holds significant importance for real-world applications. To assess the capability of output control, in this section, we determine the output length for each level based on the statistical results of the MAG dataset. For numerous practical applications, it is an unfavorable occurrence when a model generates an excessive number of labels.

From Table~\ref{tbl:controlling}, our method demonstrates strong performance in both categories. In the Paper category, it achieves counts of 0.98 at Level 0 and 1.86 at Level 1, closely aligning with MAG and significantly outperforming SGM and HECTOR. In the Patent category, "Ours" scores 1.16 at Level 0 and 1.99 at Level 1, again surpassing SGM and HECTOR. Although the Level 2 count of 4.64 is slightly above MAG's 4.19, it remains significantly better than other methods, underscoring the reliability of our approach.

\ignore{
\begin{table}[!t]
\setlength{\abovecaptionskip}{3pt}
\centering
\small
\setlength{\tabcolsep}{4pt} 
\resizebox{\linewidth}{!}{
\begin{tabular}{l|ccc|ccc}
\hline
\multirow{3}{*}{\bf Model} & \multicolumn{6}{c}{\bf MAG-Patent} \\
\cline{2-7}
& \multicolumn{3}{c|}{\bf 0-1} & \multicolumn{3}{c}{\bf 0-5} \\
& m-F & m-P & m-R & m-F & m-P & m-R  \\
\hline
\#1~~G-HXMC & 64.30 & 68.34 & 60.71 & 57.73 & 66.75 & 50.85 \\
\#2~$~~$ + Soft Cons & \bf 71.80 & \bf 76.06 & \bf 67.99 & \bf64.03 & \bf 68.48 & \bf 60.12 \\
\hline
\#3~~SGM & 59.81 & 64.23 & 55.96 & 53.01 & 39.82 & 79.25 \\
\#4~$~~$ + Global Emb & 60.62 & 65.03 & 56.77 & 53.50 & 58.01 & 49.64 \\
\#5~$~~$ + Soft Cons & 62.50 & 67.05 & 58.53 & 56.84 & 61.37 & 52.93 \\
\#6~~BERT & 63.02 & 67.14 & 59.38 & —— & —— & ——  \\
\#7~$~~$ + Soft Info & 70.69 & 75.19 & 66.70 & —— & —— & ——  \\
\hline
\end{tabular}
}
\caption{Results of MAG-Patent task in MicroF$_1$(\%)}
\label{tbl:main_results_patent}
\end{table}
}

\begin{table}[!ht]
\small
\centering
\resizebox{\linewidth}{!}{
\begin{tabular}{l|ccc|ccc}
\hline
\multirow{2}{*}{\bf Avg. Labels} & \multicolumn{3}{c}{\bf Paper} & \multicolumn{3}{c}{\bf Patent} \\
\cline{2-7}
&  Lv. 0  &  Lv. 1  &  Lv. 2  &  Lv. 0  &  Lv. 1  &  Lv. 2  \\
\hline
{MAG} & 0.80 & 1.59 & 3.64 & 1.09 & 2.13 & 4.19 \\ 

SGM & 1.30 & 2.98 & 6.89 & 1.35 & 2.54 & 3.08 \\
HECTOR & 2.10 & 3.33 & 5.01 & 2.48 & 2.88 & 7.08 \\
\hline
Ours& 0.98 & 1.86 & 4.03 & 1.16 & 1.99 & 4.64 \\ 
\hline
\end{tabular}
}
\caption{Comparison of label count in model output. }
\label{tbl:controlling}
\end{table}

\ignore{\begin{table}[!ht]
\centering
\begin{tabular}{l|ccc}
\hline
\bf Model &  \bf m-F & \bf m-P &\bf m-R \\
\hline
G-HXMC &  \bf 56.62 & \bf 61.87 & \bf 52.19 \\
SGM & 49.89 & 49.42 & 50.37 \\
HSG & 47.63 & 45.26 & 50.26 \\
\hline
\end{tabular}
\makeatletter\def\@captype{table}\makeatother\caption{The performance of Wiki10-31 on MicroF$_1$(\%).}
\label{tbl:wiki10-31}
\end{table}}

\subsection{Impact of Hierarchical Information}

\begin{table}[!ht]
\centering
\begin{tabular}{l|c}
\hline
\bf Label layers &  \bf F1\\
\hline
Level 0-1 & 71.80\\
\hline
Level 2-5 & 59.10\\
Level 0-5 & 64.03\\
\hline
\end{tabular}
\caption{Impact of utilizing parent label information .}
\label{tbl:dependence_impact}
\end{table}

Table~\ref{tbl:dependence_impact} compares the impact of utilizing parent label information on classification performance in the MAG-Patent dataset. The x-axis represents the number of encoder layers: "Level 0-1" refers to training data using only labels from levels 0 and 1; "Level 2-5" uses labels from levels 2 to 5; and "Level 0-5" includes all labels from levels 0 to 5, but evaluates classification performance only on levels 2–5.
The results indicate that the classification performance for "Level 0-1" is superior due to the smaller number of labels and the clearer label hierarchy, which reduces noise. Furthermore, training on labels from all levels (0–5) yields better classification results than using only levels 2–5. This suggests that high-quality hierarchical structure and dependency relationships present in labels from levels 0–1 provide meaningful improvements for classifying labels in levels 2–5. These findings underscore the importance and effectiveness of the proposed method in leveraging hierarchical label structure information.

\begin{table*}[!ht]
\setlength{\abovecaptionskip}{3pt}
\centering
\small
\resizebox{\linewidth}{!}{
\begin{tabular}{l|ccccccc|c}
\hline
\bf Model &  \bf Material & \bf Computer & \bf Chemical & \bf Engineering & \bf Biology & \bf Physics &\bf Medicine & \bf Avg.\\
\hline
Ours &  \bf 75.07 & \bf 63.53 & \bf 69.13 & \bf 59.20 & \bf 65.47 & \bf 61.97 & \bf 65.33 & \bf65.66 \\
SGM & 62.98 & 43.10 & 53.18 & 47.77 & 46.74 & 50.15 & 49.10 & 50.43 \\
$~~~~$ + PLC &  65.10 & 45.80 & 55.40 & 49.90 & 49.70 & 52.00 & 53.00 &52.99 \\
\hline
\end{tabular}
}
\caption{The performance of the MAG-Paper Level 0 classification task in MicroF$_1$(\%). Soft-cons means soft-constrained decoding.}
\label{tbl:results-mag-paper}
\end{table*}

\subsection{Improvement on MAG-Paper with PLC}
Table~\ref{tbl:results-mag-paper} presents the experimental results of tagging the MAG-Paper text chunks using only labels from 0-layer. Our methods demonstrate a stronger ability to leverage hierarchical structure information compared to baseline models, with a more significant gain achieved when applying the soft constraint mechanism. Specifically, the proposed method increases the average Micro-F1 score by 3.15 points using the soft constraint mechanism, while the benchmark model improves by 2.56 points.

\ignore{\begin{table}[ht]
\small
\resizebox{\linewidth}{!}{
\begin{minipage}[ht]{0.3\textwidth}
\centering
\begin{tabular}{l|c}
\hline
\bf Label layers &  \bf F1\\
\hline
Level 0-1 & 71.80\\
\hline
Level 2-5 & 59.10\\
Level 0-5 & 64.03\\
\hline
\end{tabular}
\caption{Impact of utilizing \\parent label information .}
\label{tbl:dependence_impact}
\end{minipage}
\begin{minipage}[ht]{0.2\textwidth}
\centering
\begin{tabular}{l|c}
\hline
\bf Dec len &  \bf F1\\
\hline
~5 & 35.21\\
10 &  64.03\\
15 & 60.21\\
\hline
\end{tabular}
\caption{Impact \\of different \\decoding length.}
\label{tbl:decode_length}
\end{minipage}
}
\end{table}
}

\ignore{Table~\ref{tbl:decode_length} illustrates the impact of adjusting the decoding length (i.e., the number of output labels) on classification performance in the MAG-Paper dataset. The results show that the best classification performance is achieved when the number of predicted labels closely matches the average number of labels in the training samples. This observation aligns with the authors' intuition. Additionally, the effect of the decoding length penalty coefficient on performance exhibits a similar trend. Due to space limitations, further discussion on this topic is omitted.}

\subsection{How Many Decoder Layers Are Sufficient?}

\begin{figure}[!ht]
\begin{center}
\includegraphics[width=2.0in]{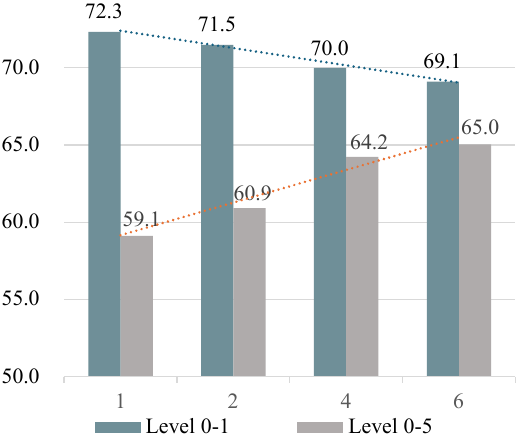}
\end{center}
\caption{Impact of different layer counts of the decoder for different level classification tasks.} 
\label{fig:encoder_layer}
\end{figure}

We also investigated the effect of encoder layer setting on different level classification tasks of the MAG-Patent dataset in Figure~\ref{fig:encoder_layer}. "Level 0-1" means that the framework output contains only the 0th and 1st level labels. "Level 0-5" means the framework generates all level labels within the taxonomy. The histogram in the figure shows that the more encoder layers used in the HMG task of 0-5 level labels, the better the performance is, which is opposite to the results of the experiment, that only our proposed framework generates only 0-1 level labels. This experiment result suggests that the more complex the dependencies between the output labels and the longer the generated sequence, the more encoder layers are needed.

\subsection{Hard Way or Soft Way?}

\begin{figure}[!ht]
\setlength{\abovecaptionskip}{0.1pt}
\setlength{\abovecaptionskip}{0pt}
\begin{center}
\includegraphics[width=2.0in]{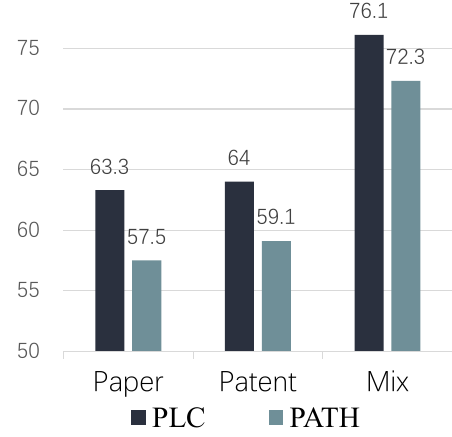}
\end{center}
\caption{Impact of different hierarchical information utilize methods.} 
\label{fig:hard_constrain}
\end{figure}

Figure~\ref{fig:hard_constrain} compares the impact of different hierarchical information utilization methods for the HMG task on the MAG-Paper dataset. The different colors refer to use our PLC mechanism or path between the labels as the guide of label generation, while the y-axis shows the Micro-F$_1$ score (\%), evaluated across varying output lengths. The hard constraint method enforces strict adherence to label hierarchy by requiring each predicted label to be a subordinate of the previous one. In contrast, the PLC method, as proposed in this paper, allows a more flexible use of label hierarchy information.

\subsection{Case Study}

A comprehensive comparison is presented, table~\ref{tbl:case_constrained} listed the resultant output labels attributed to the identical patent. This comparative analysis encompasses the utilization of two distinct methodologies for leveraging label hierarchy information. The methodologies: 1) a method devoid of both label hierarchy information and the integration of soft constraints; and 2) a contrasting approach involving the hard path constraints; 3) our proposed approach, which utilize a probability based constraint mechanism. Evidently, the application of the hard constraint methodology yields an expanded array of output labels compare with original generation model, although the outcome is marred by the conspicuous presence of labels that are overtly incongruous, such as "grease".
Conversely, the integration of PLC can precisely control the output labels of model. The labels resulting from this approach  aligning seamlessly with the specific professional domain. Evidenced by labels such as "decantation," this strategy distinctly enriches the label output with accurate and contextually relevant terminology reflective of the specific domain.

\ignore{
\subsection{Training Costs and Model Scale}

Table~\ref{tbl:time_parameters}~lists the number of parameters and training time for our model and the baseline models. Among them, the SGM model has the least number of parameters, but due to the characteristics of the recurrent neural network, its training time is the longest. The BERT model has a parameter of 109 million, and due to the lack of a decoder, the training time is the shortest under the premise of not considering the pre-training time. Our methods have balanced performance in terms of model scale, training time, and other aspects.

\begin{table}[!htbp]
\centering
\resizebox{\linewidth}{!}{
\begin{tabular}{l|c|c}
\hline
    \bf Model &  \bf Parameters (m) &  \bf Time costs (h) \\
    \hline
    G-HXMC & 66.86 & 36.5 \\
    BERT Devlin {\cite{Devlin18}} & 109.64 & \bf 30.1 \\
    SGM Yang {\cite{Yang18}} & \bf 30.99 & 50.3 \\
    \hline
\end{tabular}
}
\caption{The time costs and count of parameters of models.}
\label{tbl:time_parameters}
\end{table}

}


\begin{table}[h]
\small
\vspace{-10pt}
\begin{center}
\resizebox{\linewidth}{!}
{
\begin{tabular}{l|l}
\hline
Title & Method for decontaminating waste semi-synthetic or synthetic mineral oils \\
Abstract& c) adding to the oily phase a coagulating agent; \\
&d) adding to the oily phase barium hydroxide in water to precipitate \\
&the sulphate and phosphate ions in the form of barium sulphate and phosphate …… \\
& heating at 60 DEG C, stirring and allowing to cool and \textcolor{blue}{decanting}, …… \\
\hline
G-HXMC&…… lubricant
 $|$ petroleum \\
$~~~~$ + Hard &……  Barium hydroxide $|$ \textcolor{red}{grease} $|$ Car \\
$~~~~$ + PLC &……  Barium hydroxide $|$ Mineral oil $|$ \textcolor{blue}{decantation} \\
\hline
\end{tabular}
}
\end{center}
\vspace{-10pt}
\caption{Case study of utilizing methods of hierarchical information.} \label{tbl:zh_en}
\label{tbl:case_constrained}
\vspace{-16pt}
\end{table}

\section{Conclusion}
In this paper, we present a novel methodology that transfer Hierarchical Multi-Label Classification task to hierarchical multi-label generation task, enhanced by a refined soft constraint mechanism. This innovative approach is designed to tackle the complex challenges arising from hierarchical label relationships in HMG tasks, necessitating a context-aware and nuanced strategy.

The effectiveness of our approach is empirically validated through a comprehensive series of carefully designed experiments. To highlight the importance of our refined soft constraint mechanism, we conducted focused ablation studies. These experiments demonstrated the mechanism's crucial role in enhancing the learning process. Our results confirm that this mechanism effectively captures and leverages the hierarchical structures essential to extreme multi-label systems.

\section{Limitation}
While our approach for Hierarchical Extreme Multi-Label Generation (HMG) demonstrates significant advancements, there are limitations:

\begin{itemize}
    \item \textbf{Generalization:} The model is trained and evaluated on specific datasets, which may limit its ability to generalize to new domains with different hierarchies. This highlights the need for domain adaptation techniques to preserve performance in diverse contexts.
    \item \textbf{Domain-Specific Training:} The model's reliance on domain-specific data necessitates extensive retraining or fine-tuning when applied to new domains, which can be resource-intensive for users with limited computational resources.

    \ignore{\item \textbf{Label Noise Sensitivity:} The model's performance can be affected by noise in the label data. Errors or inconsistencies in label hierarchies can propagate through the classification process, impacting accuracy. Developing methods to mitigate the impact of label noise would be beneficial.
    
    \item \textbf{Scalability to Larger Taxonomies:} While effective for the datasets used, scaling the model to handle even larger and more complex taxonomies remains a challenge. Future work could explore more efficient architectures or training methods to address this.
    }
\end{itemize}

Future research could explore robust domain adaptation methods, efficient fine-tuning techniques, and improved strategies for handling label noise to enhance the model's applicability.

\ignore{
\section*{Ethical Statement}

There are no ethical issues.

}

\ignore{\section*{Acknowledgments}

The preparation of these instructions and the \LaTeX{} and Bib\TeX{}
files that implement them was supported by Schlumberger Palo Alto
Research, AT\&T Bell Laboratories, and Morgan Kaufmann Publishers.
Preparation of the Microsoft Word file was supported by IJCAI.  An
early version of this document was created by Shirley Jowell and Peter
F. Patel-Schneider.  It was subsequently modified by Jennifer
Ballentine, Thomas Dean, Bernhard Nebel, Daniel Pagenstecher,
Kurt Steinkraus, Toby Walsh, Carles Sierra, Marc Pujol-Gonzalez,
Francisco Cruz-Mencia and Edith Elkind.}

\bibliographystyle{named}
\bibliography{ijcai25}

\clearpage
\appendix
\section{Appendix}
\label{sec:appendix}

\subsection{Impact of Hierarchical Information utilization}

\begin{table}[h]
\setlength{\abovecaptionskip}{0.1pt}
    \centering
    \begin{tabular}[h]{l|c}
    \hline
    \bf Model &  \bf Labels\\
    \hline
    G-HXMC & 7.07 \\
    $~~~~$ + Hier-Info & \bf 7.75 \\
    SGM & 5.20 \\
    $~~~~$ + Hier-Info & 5.51 \\
    \hline
    \end{tabular}
    \caption{The length of output of our proposed model.}
    \label{tbl:layer_dependiences}
\end{table}

In this paper, the ability of our model to utilize hierarchical structure information is evaluated by observing the length of the output label sequence before and after adding structure information for different models in Table~\ref{tbl:layer_dependiences}. This analysis is based on the MAG-Paper dataset. The G-HXMC model's average increase in label count was 0.7 after adding label top-down information, while that of SGM is 0.3. Since the output label count of the SGM model was also relatively low before adding label hierarchical structure information, we speculate that the LSTM's disadvantage in encoding long sequences restricts the SGM's use of label hierarchical information and dependency relationships. As shown in Table~\ref{tbl:dataset}, the average number of labels per text in the dataset is not less than 8, and the labels have 6 levels. The SGM model's output label count means that on average, there is less than one label per level, and the output label count is relatively low, which undoubtedly reflects a negative impact on the SGM model's classification performance.

\end{document}